\documentclass[a4paper]{article}

\usepackage{INTERSPEECH2021}
\usepackage{booktabs}

\usepackage{hyperref}
\hypersetup{
    colorlinks=true,
    linkcolor=blue,
    filecolor=magenta,      
    urlcolor=cyan,
    pdftitle={Overleaf Example},
    pdfpagemode=FullScreen,
    }

\title{Skit-S2I: An Indian Accented Speech to Intent dataset}
\name{Shangeth Rajaa, Swaraj Dalmia, Kumarmanas Nethil}
\address{
  Skit.ai\\
  Bengaluru, India}
\email{shangeth.rajaa@skit.ai}

\begin{document}

\maketitle
\begin{abstract}
Conventional conversation assistants extract text transcripts from the speech signal using automatic speech recognition (ASR) and then predict intent from the transcriptions. Using end-to-end spoken language understanding (SLU), the intents of the speaker are predicted directly from the speech signal without requiring intermediate text transcripts. As a result, the model can optimize directly for intent classification and avoid cascading errors from ASR. The end-to-end SLU system also helps in reducing the latency of the intent prediction model. Although many datasets are available publicly for text-to-intent tasks, the availability of labeled speech-to-intent datasets is limited, and there are no datasets available in the Indian accent. In this paper, we release the Skit-S2I dataset, the first publicly available Indian-accented SLU dataset in the banking domain in a conversational tonality. We experiment with multiple baselines, compare different pretrained speech encoder's representations, and find that SSL pretrained representations perform slightly better than ASR pretrained representations lacking prosodic features for speech-to-intent classification. The dataset and baseline code is available at \url{https://github.com/skit-ai/speech-to-intent-dataset}
  
\end{abstract}

\noindent\textbf{Index Terms}: spoken language understanding, speech to intent, voice assistant, transfer learning

\section{Introduction}
Earlier intent classification pipeline systems transcribe the speech signal with an ASR model and then train a Natual Language Understanding(NLU) model to predict the intents using the text transcripts. These pipeline methods are prone to error propagation due to ASR transcript errors. The text-to-intent NLU models only make use of the semantic meaning of the speech and ignore other paralinguistic information in the speech which could convey the speaker's intention. The ASR and NLU models are optimized for different metrics separately and may not be the best approach for the intent classification task. This pipeline method also increases the computational requirement and latency of intent classification. Building this pipeline method is also complex as it involves training and testing two different models and the collection of both labeled corpus for training. 

End-to-End SLU methods attempt to predict the speaker's intent directly from the speech signal with a single model without any ASR transcripts. In contrast to cascaded pipeline methods, SLU methods also take advantage of other acoustic features in speech signals, including prosody. The SLU model is optimized for the target metric of intent classification, making training much simpler. SLU models also have the advantage of lesser computational requirements and better latency during deployment. The lack of a large audio dataset with SLU labels tagged still challenges SLU research. There is a need for an SLU dataset across languages, accents, and multiple domains to make the SLU models robust. In this paper, we release the Skit-S2I dataset, the first Indian-accented speech corpus for intent classification tasks in the banking domain recorded via telephony in a conversational tone. We performed an experimental comparison of cascaded pipeline methods and SLU models based on pretrained speech encoders. Also, we attempted to analyze how different pretraining methods and speech features with and without prosody information affect the performance of the SLU intent classification model. We also tried to diagnose the errors in the dataset with datamaps \cite{swayamdipta2020dataset}.

\begin{figure}[!ht]
    \centering
    \includegraphics[width=0.5\textwidth]{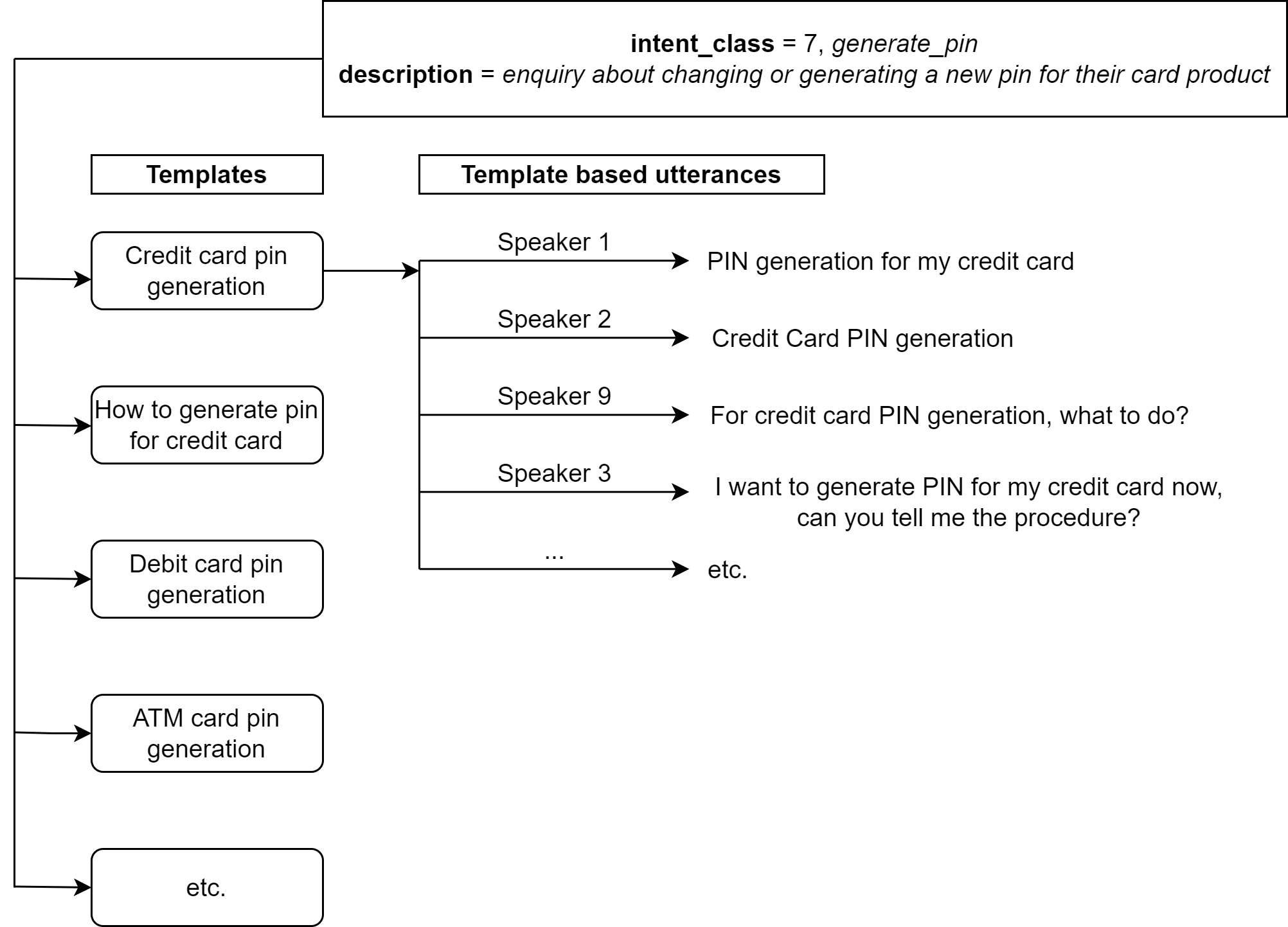}
    \caption{Example of template-based utterances for an intent class}
    \label{fig:dataset_description}
\end{figure}

\section{Related Work}
Air Travel Information System (ATIS) \cite{hemphill-etal-1990-atis} is an audio dataset with semantic labels related to air travel planning, but it is private and expensive to obtain. Snips SLU Dataset has both English and French languages, with only 2.9k and 1.2k samples in the voice assistant domain. Many previous studies have used transfer learning \cite{tomashenko19_interspeech} or acoustic model pretraining \cite{lugosch19_interspeech} for intent classification and slot prediction on smaller SLU datasets. Recently there have been efforts in curating larger datasets for end-to-end SLU tasks like FSC \cite{lugosch2019speech}, SLURP \cite{https://doi.org/10.48550/arxiv.2011.13205} and STOP \cite{https://doi.org/10.48550/arxiv.2207.10643}. However, these datasets are all based on the personal assistant domain, and no SLU datasets are available for the Indian-accented English or banking domain. In this paper, we introduce S2IDataset, which contains an Indian accented speech corpus for end-to-end SLU with multiple speakers recorded over the telephony in a conversational tonality.

\section{Skit-S2I Dataset}

\begin{table}[]
\centering
\begin{tabular}{@{}llll@{}}
\toprule
Split  & \# utterance & \# hours \\ \midrule
Train & 10445 & 12.2 \\
Test & 1400 &  1.6 \\ \midrule
Total & 11845 &  13.8\\ \bottomrule
\end{tabular}
\caption{Data split of S2IDataset}
\label{tab:data_split}
\end{table}

\subsection{Data Collection}
The Skit-S2I dataset was collected to develop voice assistants in the banking domain. The dataset consists of 14 coarse-grained intents across multiple banking-related tasks. Multiple templates have been generated for each intent class to cover possible variations in human speech. The audio utterances recorded by the speakers are spontaneously spoken based on these templates with variations. The dataset also provides descriptions for each intent. Figure \ref{fig:dataset_description} shows an example of multiple templates for intent and the variations of the speaker's utterances. The average number of templates per intent is around 12. The audio utterances were recorded over telephone calls, making channel noise possible. Background noise will not be as prevalent as in real-world scenarios as the speakers made telephone calls in a semi-controlled environment. The audio signals recorded are of 8 kHz sampling rate and 16 bit. 

\subsection{Dataset Statistics}
A total of 11 speakers contributed to the dataset, including eight females and three males. The speakers are all native Indians with different native languages from different parts of the country. There are 11845 samples in the dataset, and they are divided into train and test sets. The train set contains 10445 samples for training the model, while the test set contains 1400 samples for evaluation. Table \ref{tab:data_split} gives the data split of the Skit-S2I dataset. The train and test split have around 650 and 100 samples per intent, respectively, and each intent has data from all the speakers. The speakers are stratified across the train and test set independently for each intent. The dataset also includes anonymized speaker information such as gender, native language, languages spoken by speakers, and places lived by speakers across Indian states. Table \ref{tab:dataset_statistics} compares the statistics of different open SLU datasets.

\begin{table}[!ht]
\centering
\begin{tabular}{@{}lllll@{}}
\toprule
 & S2I & SLURP & FSC & SNIPS \\ \midrule
Domain & Banking & Assistant & Assistant & Assistant \\
Speaker & 11 & 177 & 97 & 69 \\
Audio Files & 11.8k & 72k & 30k & 5.8k \\
Duration{[}hrs{]} & 13.8 & 58 & 19 & 5.5 \\
Avg. length{[}s{]} & 4.2 & 2.9 & 2.3 & 3.4 \\ 
Intents & 14 & 91 & 31 & 7 \\ \bottomrule
\end{tabular}
\caption{Statistics of the audio corpus of different SLU datasets}
\label{tab:dataset_statistics}
\end{table}

\section{Experiments}

We experimented with several baselines of both cascaded pipeline(ASR+NLU) and end-to-end SLU models on the Skit-S2I dataset. For the cascaded pipeline, we extract the ASR transcripts using the wav2vec2 \cite{baevski2020wav2vec} model finetuned on commonvoice dataset \cite{speechbrain} and whisper small model \cite{radford2022robust}. Then we trained the pretrained XLMR \cite{conneau2019unsupervised} model using the ASR transcripts to predict the intents. 

For the end-to-end SLU models, we experimented with four different pretrained speech encoders. The intents are predicted with a linear classifier after average pooling encoded speech representations. We used the wav2vec2 \cite{baevski2020wav2vec} and Hubert \cite{hsu2021hubert} models pretrained on the self-supervised learning tasks. We also experimented with two sizes of the whisper encoder, base and small. All SLU models encode the audio with the encoders of the pretrained models. The models are finetuned with a learning rate of $1e^{-5}$ with Adam Optimizer.

\section{Results and Analysis}
Table \ref{tab:baselines-results} shows the cascaded pipeline and end-to-end SLU baseline results. The XLMR NLU model trained on different ASR transcripts gives different test accuracy and F1 scores, as the error in the ASR-predicted transcripts can affect the performance of the NLU model, and these models are not optimized together for intent classification.

\setlength{\tabcolsep}{3pt}
\begin{table}[]
\centering
\begin{tabular}{@{}lllll@{}}
\toprule
\textbf{Method} & \textbf{Model}         & \textbf{Accuracy} & \textbf{F1} \\ \midrule
Pipeline        & Wav2vec2-ASR + XLMR  & 0.923             & 0.924       \\
Pipeline        & Whisper-ASR + XLMR     & 0.933             & 0.935       \\ \midrule
E2E SLU         & Wav2vec2-SLU          & 0.953             & 0.953       \\
E2E SLU         & Hubert-SLU            & 0.955             & 0.956       \\
E2E SLU         & Whisper(base.en)-SLU  & 0.946             & 0.946       \\
E2E SLU         & Whisper(small.en)-SLU & \textbf{0.956}             & \textbf{0.957}       \\ \bottomrule
\end{tabular}
\caption{ Results for the baseline models on Skit-S2I dataset.}
\label{tab:baselines-results}
\end{table}

The Hubert-based end-to-end SLU model performed better than the wav2vec2 model, and both the whisper models outperformed Hubert and wav2vec models. In the pipeline and end-to-end SLU baselines, the whisper model outperforms the other baselines, as the whisper model is more robust and performs better in zero-shot tasks than other models. Table \ref{tab:baseline-size} compares the number of parameters of different baseline models. The pipeline baselines have the largest number of parameters as the pipeline depends on two different models, but the end-to-end SLU models are much faster with lesser computation. Whisper-based SLU models are the fastest with the lowest number of parameters as the model input is Mel-Spectrograms which requires much lesser computation than wav2vec2 or Hubert models, which input raw waveform. 

\begin{table}[]
\centering
\begin{tabular}{@{}llll@{}}
\toprule
Method   & Model                 & \# params    \\ \midrule
Pipeline & Wav2vec2-ASR + XLMR   & 315M + 278M                  \\
Pipeline & Whisper-ASR + XLMR    & 244M + 278M                  \\ \midrule
E2E SLU  & Wav2vec2-SLU          & 313M                         \\
E2E SLU  & Hubert-SLU            & 313M                         \\
E2E SLU  & Whisper(base.en)-SLU  & 19.8M                        \\
E2E SLU  & Whisper(small.en)-SLU & 87M                         \\ \bottomrule
\end{tabular}
\caption{Number of parameters for each baseline model}
\label{tab:baseline-size}
\end{table}

Most end-to-end SLU methods use the pretrained ASR features \cite{lugosch2019speech} for intent classification or use distillation methods \cite{jiang2021knowledge} to learn features from pretrained text encoders. The ASR representations will force the model to only use the linguistic/semantic information in the speech signal and ignore other important information, such as prosody. Stressing of different syllables of a word can lead to different meanings, and the overall intonation contour contributes to the speaker's intention, so prosodic features can also help improve the intent classification task. 

We compared the SSL pretrained, and ASR finetuned versions of both wav2vec2 and Hubert models for the intent classification tasks in Table \ref{tab:sslvsasr}. The SSL pretrained models slightly improve the test metrics than the ASR finetuned models. \cite{wang2021fine} also shows that SSL-trained wav2vec and Hubert models can generalize well on tasks related to learning prosody and semantic features from a speech signal, and ASR finetuning of these models lead to a loss of prosodic information in the learned representations. Based on the above results and observations, we can hypothesize that the performance of the SSL pretrained model is slightly better than ASR finetuned model, as the representation of the SSL model contains prosodic information.

\begin{table}[ht]
\centering
\begin{tabular}{@{}llll@{}}
\toprule
\textbf{Model} & \textbf{Pretraining} & \textbf{Accuracy} & \textbf{F1} \\ \midrule
Wav2vec2 & SSL & \textbf{0.953} & \textbf{0.953} \\
 & ASR-finetuned & 0.952 & 0.952 \\ \midrule
Hubert & SSL & \textbf{0.955} & \textbf{0.956} \\
 & ASR-finetuned & 0.95 & 0.951 \\ \bottomrule
\end{tabular}
\caption{Results of SSL and ASR pertained representations for intent classification}
\label{tab:sslvsasr}
\end{table}

We also trained the whisper(small.en) SLU model on the FSC and SLURP datasets to analyze the dataset as a benchmark. From Table \ref{tab:benchmark}, the whisper SLU model got an accuracy score of 0.996 on the FSC dataset and 0.765 on the SLURP dataset. The SLURP and Skit-S2I datasets are better intent classification benchmarks than FSC to compare the SLU models, as FSC test scores are much higher with simple baselines.

\begin{table}[ht]
\centering
\begin{tabular}{@{}lll@{}}
\toprule
\textbf{Dataset} & \textbf{Accuracy} & \textbf{F1} \\ \midrule
S2I & 0.956 & 0.957 \\
FSC & 0.996 & 0.996 \\
SLURP & 0.765 & 0.760 \\ \bottomrule
\end{tabular}
\caption{Results of the whisper SLU model on SLU datasets}
\label{tab:benchmark}
\end{table}

\begin{figure}[!ht]
    \centering
    \includegraphics[width=0.5\textwidth]{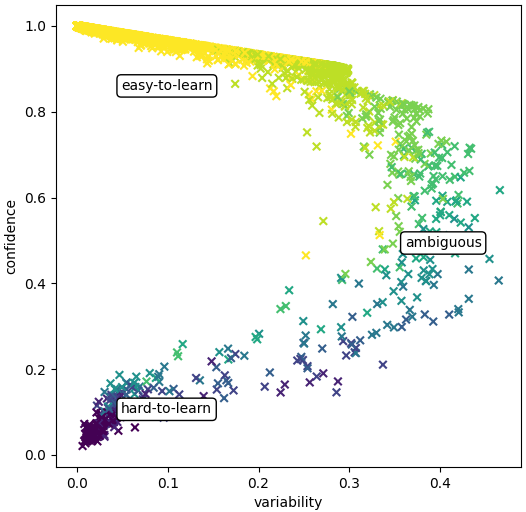}
    \caption{Datamaps Analysis of Skit-S2I dataset}
    \label{fig:dataset_cartography}
\end{figure}

\begin{figure}[!ht]
    \centering
    \includegraphics[width=0.23\textwidth]{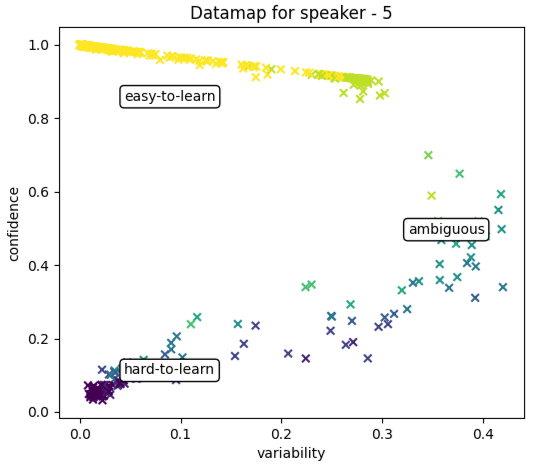}\hfill
    \includegraphics[width=0.23\textwidth]{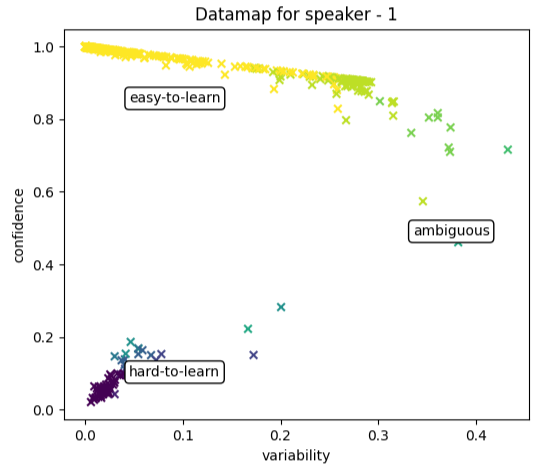}
    \\[\smallskipamount]
    \includegraphics[width=0.23\textwidth]{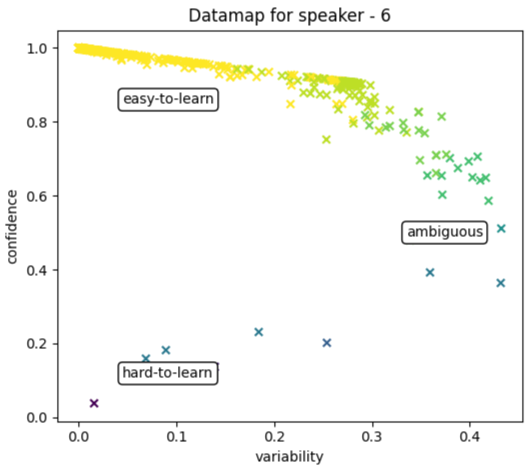}\hfill
    \includegraphics[width=0.23\textwidth]{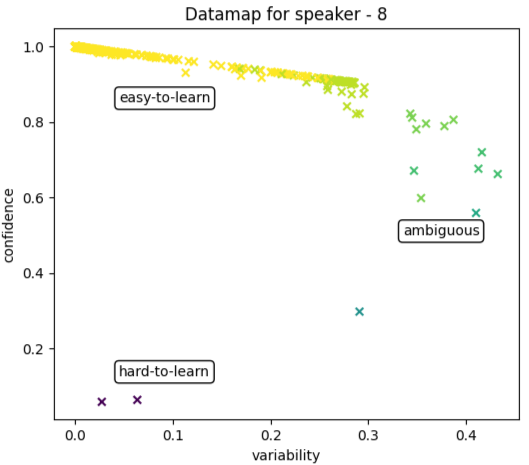}

    \caption{Datamaps for Skit-S2I dataset for speakers 1,5,6 and 8.}
    \label{fig:dataset_cartography_speakers}
\end{figure}

\begin{figure}[!ht]
    \centering
    \includegraphics[width=0.5\textwidth]{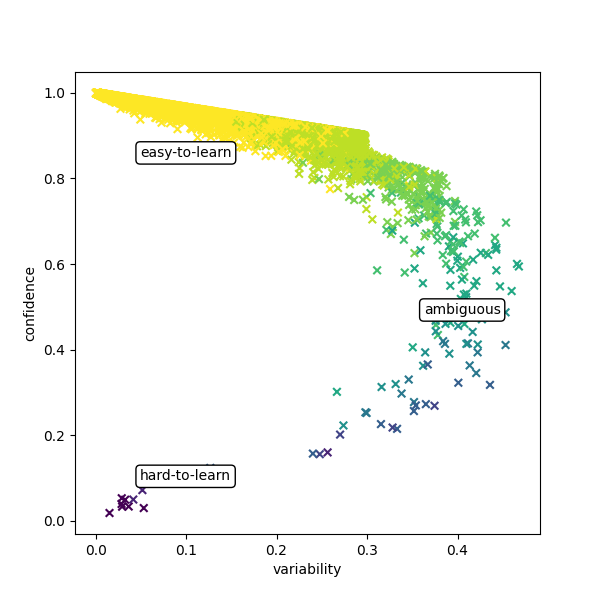}
    \caption{Datamaps Analysis of FSC dataset}
    \label{fig:dataset_cartography_fsc}
\end{figure}

\begin{figure}[!ht]
    \centering
    \includegraphics[width=0.5\textwidth]{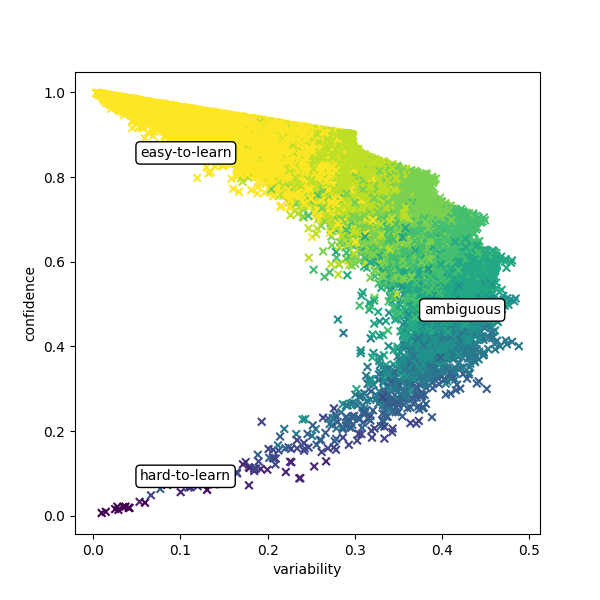}
    \caption{Datamaps Analysis of SLURP dataset}
    \label{fig:dataset_cartography_slurp}
\end{figure}

\section{Dataset Analysis}
We performed the dataset cartography \cite{swayamdipta2020dataset} analysis on the S2IDataset, to assess the quality and diagnose the dataset. Dataset Cartography leverage the training dynamics of a model trained on a dataset to create the datamaps, which split the dataset into three distinct regions, easy-to-learn, ambiguous and hard-to-learn. The ambiguous samples in the dataset contribute to the out-of-distribution generalization, the easy-to-learn samples play an essential role in the optimization, and hard-to-learn samples often correspond to labeling errors. Figure \ref{fig:dataset_cartography} shows the datamap generated by training the dataset with the Whisper(small.en) model. After generating the datamaps, we found that a few of the data points in the hard-to-learn samples had label noise, speechless audio files, and short speech utterances, which may be because of telephony noise. Figures \ref{fig:dataset_cartography_slurp} and \ref{fig:dataset_cartography_slurp} show the generated datamaps for the FSC and SLURP datasets, respectively. Most samples in the FSC dataset were in the easy-to-learn region, whereas the SLURP dataset has a good split of data samples across all the regions. Thus SLURP dataset can act as a better benchmark than FSC.

We also generated the datamaps for each speaker in the dataset to identify which speaker's audio samples had the above issues. We examined the generated datamaps for all the speakers and found that most of the dataset errors were samples from only a few speakers. Figure \ref{fig:dataset_cartography_speakers} shows the datamaps generated for speakers 1, 5, 6, and 8. Speakers 6,8 had very few hard-to-learn samples, whereas speakers 1 and 5 had many hard-to-learn samples and data errors. Not all hard-to-learn samples are data errors, but they have a high chance of being an error. So we manually check all the hard-to-learn samples and change/remove them to make the dataset error-free. Based on the dataset cartography analysis, we will remove the errors in the future version of the S2IDataset dataset.

\section{Conclusions}
In this work, we released the Skit-S2I dataset, the first Indian-accented SLU dataset for the banking domain. We trained and compared multiple cascaded pipeline-based and end-to-end SLU baselines. We also compared the performance of different models and found that SSL representations perform better than ASR, as SSL representations contain prosodic information. We also compared the performance of the whisper-based intent classification model on FSC and SLURP datasets and found that the baseline model was able to achieve very good performance on the FSC evaluation. Skit-S2I and SLURP are better benchmarks than the FSC dataset.

\bibliographystyle{IEEEtran}

\bibliography{main}

\end{document}